\documentclass{article}
\usepackage{spconf,amsmath,graphicx}
\usepackage{booktabs}
\usepackage{makecell}
\usepackage[disable]{todonotes}
\usepackage{caption}
\usepackage{url}

\title{Detecting Check-Worthy Claims in Political Debates, Speeches,\\ and Interviews Using Audio Data}
\name{Petar Ivanov,$^1$ Ivan Koychev,$^1$ Momchil Hardalov,\thanks{$^*$Work done while Momchil Hardalov was at the Sofia University, prior to joining Amazon.}$^2$ Preslav Nakov$^3$}
\address{
  $^1$FMI, Sofia University ``St. Kliment Ohridski'' \quad $^2$AWS AI Labs \\ $^3$ Mohamed bin Zayed University of Artifcial Intelligence\\
{peturoi@uni-sofia.bg, koychev@fmi.uni-sofia.bg, momchilh@amazon.com, preslav.nakov@mbzuai.ac.ae}
}

\begin{document}

\maketitle
 
\begin{abstract}

Developing tools to automatically detect check-worthy claims in political debates and speeches can greatly help moderators of debates, journalists, and fact-checkers. While previous work on this problem has focused exclusively on the text modality, here we explore the utility of the audio modality as an additional input. We create a new multimodal dataset (text and audio in English) containing 48 hours of speech from past political debates in the USA. We then experimentally demonstrate that, in the case of multiple speakers, adding the audio modality yields sizable improvements over using the text modality alone; moreover, an audio-only model could outperform a text-only one for a single speaker. With the aim to enable future research, we make all our data and code publicly available at \url{https://github.com/petar-iv/audio-checkworthiness-detection}.
\end{abstract}

\textbf{\emph{Index Terms}:} Check-Worthiness, Fact-Checking, Fake News, Misinformation, Disinformation, Political Debates, Multimodality.

\section{Introduction}

Manual fact-checking is the most important and credible way to fight mis/disinformation. Yet, it is very tedious and time-consuming, and thus it is important to prioritize what to fact-check, i.e., to estimate the check-worthiness of the claims~\cite{10.1145/3395046,ijcai2020p672,hardalov-etal-2022-survey}. Here, we focus on political debates and speeches. An example is shown in Table~\ref{tab:text-snippet}, where each sentence in  political debate is annotated with the person who said it and whether it is check-worthy. Following previous work, we consider as check-worthy sentences that contain claims that have been fact-checked by human fact-checkers \cite{check-that-2021}.

\begin{table}[th]
  \caption{Snippet of the textual transcript of a political debate.}
  \label{tab:text-snippet}
  \centering
  \setlength{\tabcolsep}{3pt}
  \begin{tabular}{c c c c}
    \toprule
    \textbf{Line} & \textbf{Speaker} & \textbf{Sentence} & \textbf{Claim} \\
    \midrule
    146 & PENCE & \makecell{But Hillary Clinton and Tim Kaine \\ want to build on Obamacare.} & 0 \\
    \hline
    147 & PENCE & \makecell{They want to expand it into \\ a single-payer program.} & 1 \\
    \hline
    842 & KAINE & \makecell{The Clinton Foundation is one of the \\ highest-rated charities in the world.} & 0 \\
    \hline
    843 & KAINE & \makecell{It provides AIDS drugs to about \\ 11.5 million people.} & 1 \\
    \bottomrule
  \end{tabular}
\end{table}

Our contributions can be summarized as follows: (\emph{i}) a new multimodal dataset (text and audio) for detecting check-worthy claims, (\emph{ii}) a novel framework that combines the text and the speech modalities, (\emph{iii}) evaluation and comparison of current state-of-the-art textual and audio models on our multimodal dataset.

\section{Related Work}

\textbf{Detecting check-worthy claims.}
The CheckThat! lab is a competition held annually, featuring tasks related to misinformation and fake news. One of them is detecting check-worthy claims in the transcripts of political debates. The best system \cite{check-that-2019-winner} in the 2019 challenge \cite{check-that-2019} relied on specially trained word2vec embeddings~\cite{DBLP:journals/corr/abs-1301-3781}, syntactic dependencies within a sentence, an LSTM neural network, as well as soft labeling. In 2020 \cite{check-that-2020}, the winning system used 6B-100D GloVe embeddings and a bidirectional LSTM \cite{check-that-2020-winner}. In 2021 \cite{check-that-2021}, the best system \cite{check-that-2021-winner} used distilRoBERTa~\cite{sanh2019distilbert} (with an extra dropout and classification layers) fine-tuned on a hate speech dataset.

\noindent\textbf{Using audio data in similar domains.} To the best of our knowledge, we are the first to use audio data for detecting check-worthy claims. Thus, we will discuss work that uses speech in similar domains. 
The goal in \cite{kopev2019detecting} is to classify a factual claim as true, half-true, or false. The authors built a multimodal dataset (text + audio) containing 33 minutes of speech. The experimental results showed that using both textual and audio input helps over using text only. 
The task in \cite{Dinkov2019} was to predict the leading political ideology (left, center, or right) of a news mediuim's YouTube channel using text and audio input. The results demonstrated the utility of the audio signal when added to text. Finally, an area related to check-worthiness is deception detection in audio/video signals~\cite{FathimaBareeda_2021,8695348,talaat2023explainable,9025340}.

\section{Data}

Our new multimodal dataset combines the textual transcripts of political debates, speeches, and interviews (referred to as \emph{events}) with their original audio recordings. 
It is an augmentation of the dataset for the 2021 CheckThat! lab~\cite{check-that-2021}, Task 1b. Table~\ref{tab:data} shows the distribution of the utterances in the original and in our augmented dataset. We can see that our new dataset is a bit smaller in size as some recordings could not be found and some were only partial.

\begin{table}[h]
  \caption{Comparison between the original CheckThat!'21 dataset and our multimodal one.}
  \label{tab:data}
  \centering
  \setlength{\tabcolsep}{3pt}
  \begin{tabular}{lrr}
    \toprule
    \textbf{} & \textbf{CheckThat!'21} & \textbf{Our Dataset} \\
    \textbf{Modality} & {Text Only} & {Text + Audio} \\
    \midrule
    \multicolumn{3}{l}{\bf Train} \\
    \# events & 40 & 38 \\
    \# sentences & 42,033 & 28,715 \\
    \# check-worthy claims & 429 & 417 \\
    \midrule
    \multicolumn{3}{l}{\bf Dev} \\
    \# events & 9 & 7 \\
    \# sentences & 3,586 & 1,896 \\
    \# check-worthy claims & 69 & 40 \\
    \midrule
    \multicolumn{3}{l}{\bf Test} \\
    \# events & 8 & 8 \\
    \# sentences & 5,300 & 3,878 \\
    \# check-worthy claims & 298 & 291 \\
    \midrule
    \multicolumn{3}{l}{\bf All} \\
    \# events & 57 & 53 \\
    \# sentences & 50,919 & 34,489 \\
    \# check-worthy claims & 796 & 748 \\
    \bottomrule
  \end{tabular}
\end{table}

\noindent\textbf{Data Distribution.} We can see in Table~\ref{tab:data} that the data distribution is extremely skewed: the check-worthy claims are around 2\% of all sentences. Thus, we prepared three variants of the training dataset based on oversampling and undersampling: (\emph{i})~upsample the check-worthy statements 15 times (referred to as \emph{x15}), (\emph{ii})~upsample them 30 times (referred to as \emph{x30}), and (\emph{iii})~remove random non-check-worthy sentences until their number becomes equal to the number of the check-worthy ones (referred to as \emph{1:1}). We summarize these different variations of the training dataset in Table~\ref{tab:train-dataset-variations}.

\begin{table}[t]
  \caption{Statistics about variations of the training dataset with different sampling ratios.}
  \label{tab:train-dataset-variations}
  \centering
  \setlength{\tabcolsep}{3pt}
  \begin{tabular}{lrrrr}
    \toprule
    \textbf{} & \textbf{Original} & \textbf{x15} & \textbf{x30} & \textbf{1:1} \\
    \midrule
    \# non-check-worthy & 28,298 & 28,298 & 28,298 & 417 \\
    \# check-worthy claims & 417 & 6,672 & 12,927 & 417 \\
    \midrule
    Check-worthy claims & 1.5\% & 19.1\% & 31.4\% & 50.0\% \\
    \bottomrule
  \end{tabular}
\end{table}

\noindent\textbf{Single-Speaker Variant.} The dataset includes utterances from multiple speakers, which have different accents and pronunciations. We propose a single-speaker setup, leaving aside the speech specifics of the different speakers. Those with the most check-worthy claims in the training dataset are Donald Trump (51\% of all check-worthy), Hillary Clinton (17\%), and Bernard Sanders (7\%). We create a separate dataset with Donald Trump's utterances only. See Table~\ref{tab:trump-sentences} for detailed statistics.

\begin{table}[t]
  \caption{Statistics about our single-speaker dataset.}
  \label{tab:trump-sentences}
  \centering
  \setlength{\tabcolsep}{3pt}
  \begin{tabular}{l r r r r}
    \toprule
    \textbf{} & \textbf{Train} & \textbf{Dev} & \textbf{Test} & \textbf{All} \\
    \midrule
    \# sentences & 8,191 & 1,650 & 3,489 & 13,330 \\
    \# check-worthy claims & 213 & 39 & 278 & 530 \\
    \midrule
    Check-worthy claims  & 2.6\% & 2.4\% & 8.0\% & 4.0\% \\
    \bottomrule
  \end{tabular}
\end{table}

\noindent\textbf{Audio Noise Reduction.} We prepared a separate variant of the utterances in the audio files (the \emph{audio segments}) with reduced background noise, using \emph{noisereduce} \cite{noise-reduce-1, noise-reduce-2}.

\section{Proposed Joint Models}

\subsection{Knowledge Alignment from Text to Audio}
\label{subsec:alignment}

\begin{figure}[th]
  \centering
  \caption{Overview of the knowledge alignment mechanism between the textual representations and the audio.}
  \includegraphics[width=\columnwidth]{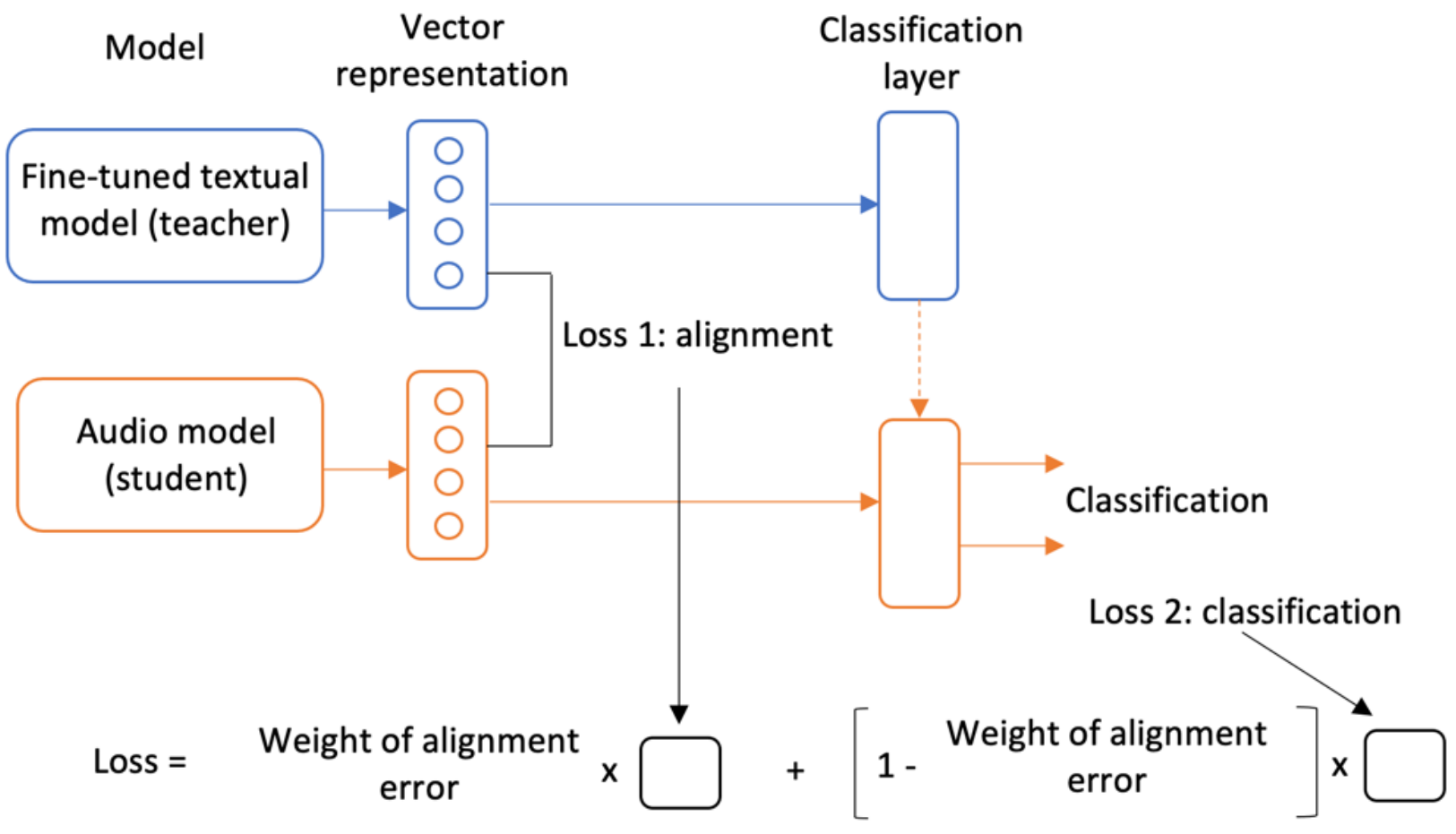}
  \label{fig:alignment}
\end{figure}

Large language models (LLMs) have demonstrated impressive results for many text-based tasks~\cite{bert}. They are also the backbone of current state-of-the-art systems for discovering check-worthy claims~\cite{check-that-2021-winner}. However, as they take text as input, we need to perform an additional step: transcription from audio to text, which can be costly and time-consuming. On the other hand, audio models~\cite{wav2vec2, hubert, data2vec} can produce latent representations from the audio that can remove the need for extra steps, but these are not as powerful for complex language tasks per se. That said, we hypothesize that further alignment between the task-specific representations of the audio and of the textual models can significantly improve the audio performance. 
In particular, we train an audio model to represent the input it receives in the same way that a fine-tuned textual model would represent its input in a teacher-student mode. Figure~\ref{fig:alignment} illustrates this approach. The audio model reuses the classification layer of the textual one. The training is on two tasks: vector alignment and classification. We calculate two loss values via mean squared error and cross-entropy. The final (or composite) loss is a weighted sum of these two losses using a hyper-parameter $\lambda$. The weight of the alignment error is provided as an input to the training script and the other weight is calculated accordingly.

\subsection{Ensembles}

Next, we combine representations from separately fine-tuned audio and textual models, as previous studies~\cite{kopev2019detecting, Dinkov2019} have shown that the two modalities can be complementary. 

\noindent We leverage both the audio and the textual representations by combining them using two techniques: (\emph{i})~\emph{early fusion} -- concatenating the two vector representations into a single feature vector and passing it to a classifier, and (\emph{ii})~\emph{late fusion} -- taking the confidence of each model that the sentence is check-worthy, and building a single vector with two values and then passing it to a classifier, which we need to train in this setup.

\section{Experiments}

We used Mean Average Precision (MAP) as an evaluation measure as it is also the main measure in the 2021 CheckThat! lab~\cite{check-that-2021}.

\subsection{Single-Modality Baselines}

For all pre-trained models, we use the public checkpoints from HuggingFace \cite{wolf-etal-2020-transformers}.

\noindent\textbf{Text-Only Models:} (\emph{i})~Feedforward network using the number of named entities,\footnote{Extracted using SpaCy, a total of 18 types of named entities.}(\emph{ii})~Support Vector Machine based on TF.IDF\footnote{Baseline provided by the 2021 CheckThat! lab organizers.}, and (\emph{iii})~fine-tuned BERT-base uncased~\cite{bert}.

\noindent\textbf{Audio Models:} (\emph{i})~wav2vec 2.0~\cite{wav2vec2}, (\emph{ii})~HuBERT~\cite{hubert}, and (\emph{iii})~data2vec-audio~\cite{data2vec}. All the audio models are based on the Transformer architecture~\cite{nips2017_7181:transformer}. We use their base variants.

\subsection{Experimental Setup}

We used AdamW~\cite{adamw} with a warm-up proportion of 0.1 and a weight decay of 0.02. Feed-forward neural networks have ReLU activation and a dropout after every hidden layer. We trained the models using a cross-entropy classification loss, except for the model with knowledge alignment (see Section~\ref{subsec:alignment}).
We used $\lambda$ = 0.75 as a weight for the alignment loss (and 0.25 for the classification loss) when aligning the audio representations. 

\noindent The maximum input length for the BERT-based models is 100 tokens (5\% of the samples are truncated) and for audio, we cut the segment after 8 seconds, which affects 20\% of the samples. We chose the best checkpoint on the \emph{dev}elopment dataset. We set the number of training epochs to 15. We trained HuBERT with a learning rate of $7\times10^{-5}$, and wav2vec 2.0 and data2vec-audio with $5\times10^{-5}$. 

\subsection{Experimental Results}
 
Tables~\ref{tab:results-multiple-speakers-textual},~\ref{tab:results-multiple-speakers-audio},~\ref{tab:results-multiple-speakers-aligned-audio}, and~\ref{tab:results-multiple-speakers-ensembles} show the results on the full multimodal dataset (i.e.,~with multiple speakers), and Tables~\ref{tab:results-single-speaker-textual} and~\ref{tab:results-single-speaker-audio} on the dataset with Trump's sentences (single speaker). 
We report only the best results for each combination of model, training dataset variant, and audio segments variant.

\textbf{Multiple Speakers.} In the multiple-speaker setting, we can see that simple textual baselines (Table~\ref{tab:results-multiple-speakers-textual}) achieve a MAP of 22.28 for the FNN (row 3: it uses 1 hidden layer of size 8, and a learning rate of 0.05) and 23.92 for the SVM (row 2: it uses an \emph{RBF} kernel with $\gamma=0.75$) -- that is approximately 15 points below our best model (see Table~\ref{tab:results-multiple-speakers-ensembles}, row 1).

\begin{table}[h]
  \caption{Best results with textual models over the full multimodal dataset (multiple speakers).}
  \label{tab:results-multiple-speakers-textual}
  \centering
  \small
  \begin{tabular}{l c c c}
    \toprule
    \textbf{Row} & \textbf{Model} & \textbf{\makecell{Train \\dataset}} & \textbf{MAP(test)} \\
    \midrule
    1 & BERT & 1:1 & 37.15 \\
    \hline
    2 & \makecell{SVM with \\TF.IDF} & x15 & 23.92 \\
    \hline
    3 & \makecell{Feedforward network with\\named entities count} & x15 & 22.28 \\
    \bottomrule
  \end{tabular}
\end{table}

Audio models (Table~\ref{tab:results-multiple-speakers-audio}) show improvement over the simpler textual baselines: data2vec-audio (row 3) and wav2vec 2.0 (row 2) outperform FNN, while HuBERT (row 1) is the second best among the models based on a single modality (Tables~\ref{tab:results-multiple-speakers-textual} and~\ref{tab:results-multiple-speakers-audio}) with a MAP of 25.26.

\begin{table}[h]
  \caption{Best results with audio models over the full multimodal dataset (multiple speakers).}
  \label{tab:results-multiple-speakers-audio}
  \centering
  \setlength{\tabcolsep}{3pt}
  \begin{tabular}{l c c c c}
    \toprule
    \textbf{Row} & \textbf{Model} & \textbf{\makecell{Train \\dataset}} & \textbf{\makecell{Audio \\segments}} & \textbf{MAP(test)} \\
    \midrule
    1 & HuBERT & x30 & Original & 25.26 \\
    \hline
    2 & wav2vec 2.0 & x15 & Original & 23.65 \\
    \hline
    3 & data2vec-audio & x30 & Reduced noise & 23.30 \\
    \bottomrule
  \end{tabular}
\end{table}

Next, we can see that our alignment procedure from Section~\ref{subsec:alignment} yields an improvement over training on classification only (Table~\ref{tab:results-multiple-speakers-aligned-audio}), adding more than 6 points absolute MAP to the score of wav2vec and data2vec reaching MAP of almost 30, and adding almost 3 points to the HuBERT model.

\begin{table}[h]
  \caption{Best results when using audio models with alignment over the full multimodal dataset (multiple speakers).}
  \label{tab:results-multiple-speakers-aligned-audio}
  \centering
  \setlength{\tabcolsep}{4pt}
  \begin{tabular}{l c c c c c}
    \toprule
    \textbf{Row } & \textbf{Model} & \textbf{\makecell{Train \\dataset}} & \textbf{\makecell{Audio \\segments}} & \textbf{MAP(test)} \\
    \midrule
    1 & data2vec-audio & \makecell{Without\\changes} & Original & 29.99 \\
    \hline
    2 & wav2vec 2.0 & \makecell{Without\\changes} & Original & 29.96 \\
    \hline
    3 & HuBERT & \makecell{Without\\changes} & Original & 27.87 \\
    \bottomrule
  \end{tabular}
\end{table}

Nevertheless, all these models perform much worse than plain BERT (Table~\ref{tab:results-multiple-speakers-textual}, row 1), which achieved a MAP of 37.15 (with a weight decay of 0.01 and a learning rate of $2\times10^{-5}$).

The ensembles of BERT and an audio model yielded a higher MAP score (Table~\ref{tab:results-multiple-speakers-ensembles}) than BERT alone (Table~\ref{tab:results-multiple-speakers-textual}, row 1). We used a feed-forward network with two hidden layers. The hidden layer sizes for the rows in Table~\ref{tab:results-multiple-speakers-ensembles} are (256, 64), (6, 6), (512, 256), and (6, 6), respectively. We used a learning rate of 0.001 in all cases, except for row 3, for which it was 0.0001. The respective dropouts are 0.1, 0, 0.4, and 0.

\begin{table}[h]
  \caption{Best results with text+audio ensembles over the full multimodal dataset (multiple speakers).}
  \label{tab:results-multiple-speakers-ensembles}
  \centering
  \small
  \setlength{\tabcolsep}{3pt}
  \begin{tabular}{l c c c c c}
    \toprule
    \textbf{Row} & \textbf{\makecell{Ensemble\\type}} & \textbf{Model} & \textbf{\makecell{Train\\dataset}} & \textbf{\makecell{Audio\\segments}} & \textbf{MAP(test)}  \\
    \midrule
    1 & \makecell{Early\\fusion} & \makecell{BERT \& \\ HuBERT} & \makecell{Without\\changes} & Original & 38.17 \\
    \hline
    2 & \makecell{Late\\fusion} & \makecell{BERT \& \\ HuBERT} & x15 & Original & 37.58 \\
    \hline
    3 & \makecell{Early\\fusion} & \makecell{BERT \& \\ aligned data2vec} & \makecell{Without\\changes} & Original & 37.35 \\
    \hline
    4 & \makecell{Late\\fusion} & \makecell{BERT \& \\ aligned data2vec} & x30 & Original & 37.24 \\
    \bottomrule
  \end{tabular}
\end{table}

Comparing the early fusion ensemble with HuBERT to the early fusion ensemble with aligned data2vec-audio (Table~\ref{tab:results-multiple-speakers-ensembles}, rows 1 and 3), we can see that HuBERT has a lower MAP of 25.26 (Table~\ref{tab:results-multiple-speakers-audio}, row 1) compared to the aligned data2vec-audio model, which has a score of 29.99 (Table~\ref{tab:results-multiple-speakers-aligned-audio}, row 1), but the results when used in an ensemble are inverse. The vector representations of HuBERT go through an additional projection layer, which reduces the dimensionality from 768 to 256, and thus the information is more condensed, which yields slightly better results. With data2vec-audio, all 768 values in the vector representation are considered.

\textbf{Single-Speaker Results.} In this setup, wav2vec 2.0 achieves a score (Table~\ref{tab:results-single-speaker-audio}, row 1) that is even higher than for BERT (Table~\ref{tab:results-single-speaker-textual}, row 1). The textual models use the same hyper-parameter values as in the multiple-speaker setup, with one exception: the learning rate for BERT is $1\times10^{-5}$.

\begin{table}[h!]
  \caption{Best results with textual models over the single speaker (Trump) subset of the multimodal dataset.}
  \label{tab:results-single-speaker-textual}
  \centering
  \small
  \begin{tabular}{l c c c}
    \toprule
    \textbf{Row} & \textbf{Model} & \textbf{MAP(test)} \\
    \midrule
    1 & BERT & 32.67 \\
    \hline
    2 & \makecell{SVM with \\TF.IDF} & 26.93 \\
    \hline
    3 & \makecell{Feedforward network with\\named entities count} & 21.93 \\
    \bottomrule
  \end{tabular}
\end{table}

\begin{table}[h!]
  \caption{Best results with audio models for the single-speaker subset of the multimodal dataset.}
  \label{tab:results-single-speaker-audio}
  \centering
  \begin{tabular}{l c c c}
    \toprule
    \textbf{Row} & \textbf{Model} & \textbf{\makecell{Audio \\segments}} & \textbf{MAP(test)} \\
    \midrule
    1 & wav2vec 2.0 & Reduced noise & 34.27 \\
    \hline
    2 & HuBERT & Original & 24.78 \\
    \hline
    3 & data2vec-audio & Reduced noise & 21.29 \\
    \bottomrule
  \end{tabular}
\end{table}

We further explored 100-dimensional i-vectors \cite{kopev2019detecting, Dinkov2019} (extracted with \emph{Kaldi}); \emph{ComParE 2013}, \emph{ComParE 2016}, and \emph{MFCC} features, but the results did not surpass those with the best audio models (HuBERT and aligned data2vec-audio) for multiple speakers and wav2vec 2.0 for a single speaker.

\section{Conclusion and Future Work}

We experimented with the audio modality in the task of detecting check-worthy claims in political debates, interviews, and speeches. We built a multimodal dataset (48 hours of speech), which comprises 34,489 sentences. We addressed the class imbalance in the training dataset and prepared several variants of it. We also had two variants of the audio segments. The results showed that in the case of multiple speakers, audio models yielded higher MAP scores than textual baselines. Ensembles with BERT and an audio model (best MAP 38.17) showed improvement over BERT alone (MAP 37.15). For a single speaker, an audio model achieved a better result (MAP 34.27) than BERT (MAP 32.67). 
The contributions of our work include building a new multimodal dataset, a novel framework that combines the text and the speech modalities, and experiments showing positive results.

In future work, we plan to experiment with learning-to-rank, using the context \cite{gencheva-etal-2017-context,vasileva-etal-2019-takes}, i.e.,~utterances/audio segments that come before and after the current one, addressing label imbalance via SMOTE~\cite{chawla2002smote}, using models for deception detection in audio.

\vspace{6pt}
\noindent{\bf  ACKNOWLEDGMENT} 
This work is partially funded by project UNITe BG05M2OP001-1.001-0004 by the OPSESG and by the EU - NextGenerationEU, through the National Recovery and Resilience Plan of the Republic of Bulgaria, project No BG-RRP-2.004-0008.


\bibliographystyle{IEEEbib}
\bibliography{mybib}

\end{document}